# Irony Detection in Urdu Text: A Comparative Study Using Machine Learning Models and Large Language Models

[1] Fiaz Ahmad, [2] Nisar Hussain, [2] Amna Qasim, [2] Momina Hafeez, Muhammad Usman [2] Grigori Sidorov, [2] Alexander Gelbukh

[1] The University of Central Punjab (UCP), Punjab, Pakistan,

[2] Instituto Politécnico Nacional (IPN), Centro de Investigación en Computación (CIC), Mexico,

fiaz.ahmad6917@gmail.com, nhussain2022@cic.ipn.mx, amnaq2023@cic.ipn.mx, mhafeez2025@cic.ipn.mx, usmancic21@gmail.com, sidorov@cic.ipn.mx, gelbukh@cic.ipn.mx

*Abstract*— *Ironic identification is a difficult issue in the context of NLP, particularly when we switch language with different syntactic and cultural differences. Here we aim to detect irony by translating the Ironic Corpus available for English language into the Urdu language. We next utilize ten states of the art machine learning algorithms with GloVe and Word2Vec embeddings to compare with classical methods. We also fine-tune more sophisticated transformer-based models viz., BERT, RoBERTa, LLaMA 2 (7B), LLaMA 3 (8B) and Mistral, to see the performance of large-scale models in detecting irony. On machine learning models, Gradient Boosting had the best performance with an F1-Score of 89.18%. Among the top-performing LLMs, LLaMA 3 (8B) achieved the highest performance, reaching an impressive F1-Score of 94.61%. This study confirms that integrating transliteration engines with state-of-the-art NLP models enables robust irony detection in Urdu, a historically low-resource language.*

## I. INTRODUCTION

Irony detection is widely studied in the field of Natural Language Processing (NLP) as it greatly affects the performance of sentiment analysis, content moderation and opinion mining systems. Unlike sentiment classification, which is relatively straightforward, irony detection is a more complex task that places greater emphasis on context, shifts in sentiment, and underlying meanings that are not always immediately apparent [1, 2]. Studies have tried to consider different features (linguistic, syntactic, and affective) in the attempt to improve irony detection for text, and especially with short and less formally constructed text, typical on Twitter.

Previous works also broadened the detection of irony beyond English by examining the phenomenon in other languages such as Arabic [3], Spanish varieties [4], or Dutch [5, 6]. In cross-linguistic settings, irony becomes even more complex due to differences in cultural representations and linguistic constructions at both textual and discursive levels [7]. Early methods depend on feature engineering, where handcrafted features are used to model irony, with shallow models being employed [8]; now however, transformer-based architectures, including BERT and RoBERTa, can be exploited to capture complex irony patterns. Ensemble methods and transfer learning approaches have also demonstrated success, improving the generalization capabilities of irony detectors across various domains and languages [9].

Nevertheless, there have been very few efforts toward irony detection in Urdu. The absence of a publicly available Urdu irony dataset, combined with the unique complexity of Urdu syntax and cultural humor, highlights the need for dedicated investigation. Filling in these gaps, we translate the widely used Ironic Corpus from English into Urdu and annotate it with the original annotations and accounting for linguistic nuances. This paper provides an extensive examination across ten machine learning models and several large language models (LLMs), e.g., BERT, RoBERTa, LLaMA 2, LLaMA 3, and Mistral. In this work, we attempt to address the irony detection scarcity for Urdu and show a possibility for cross-lingual resource adaptation.

## II. RELATED WORK

A variety of methods for irony detection have been addressed in recent literature, ranging from linguistic cues to deep learning techniques. For instance, while transformer-based models have been successful, issues related to bias in the mechanisms have begun to emerge as reported by [9]. Another challenging direction is multimodal irony detection including text and visual modalities [10] that underline the increasing challenges of multimodality.

Effectively detecting irony requires considering related connotations and contexts, as demonstrated by [11], which examines how model predictions vary in relation to annotators. [12] proposed the emotion feature augmentation with large language models in order to further improve the detection of ironic sentences. [13] also offered models for irony detection and use by proposing quantitative and mathematical reasoning under a zero-shot learning framework, thereby paving the way for more generalized systems.

Domain adaptation remains another important aspect in which [14] proposed ensemble methods for adapting irony detection models to other social media domains. Deep learning techniques are thoroughly surveyed by [15] in respect of the relative potential of different architectures. In non-English languages, [16] introduced a semantically informed graph neural network in the task of irony detection in Turkish.

Outside of text, [17] investigated multi-modality of acoustic and facial expression for irony detection presenting some promising interdisciplinary approaches. Text analysis of historical text has also developed, with [18] testing of irony identification in written Spanish 19th c. language using large language models.

The progress in ironic detection in Arabic was discussed by [19], with a focus on shifting from emotion AI to cognition AI behavior. [20] investigated the problem of machine translation in multilingual irony detection, and proposed approaches to improve translation settings. The perspectivism method proposed by [21] also contributed to the diversity in irony annotation and detection via the PERSEID challenge.

Further, [22] studied profiling of irony spreaders in social networks for improving user-based irony detection. Psychological and cognitive factors have been associated with the comprehension of irony, [21] who linked cognitive flexibility and trait anger with the ability to perceive and understand irony. Finally,[23] investigated the interpretation of depression by large language models, revealing recent tendencies for depression aspect.

### III. METHODOLOGY
#### 3.1. Dataset Description

The dataset we use in the present study is the "Ironic Corpus," which was previously used for irony detection in English. It contains 1,950 Reddit comments marked with the tags ironic and non-ironic. Acknowledging the paucity of irony resources in the Urdu language, we translated this dataset to Urdu using a semi-automatic process which involved the use of machine translation and human post-editing. Particular emphasis was placed on not only ensuring the accuracy of the intended meaning, but also the preservation of irony, as many idioms themselves fail to translate directly, thereby running the risk of losing any nuanced application of context or sense of irony.

The item translations were subjected to a multi-stage process of quality assurance to guarantee the semantic equivalent between the source and target languages. Other NASTI annotators fluent in English & Urdu grammar corroborated the maintenance of irony, wit & sarcasm when necessary. The final Urdu dataset keeps the same labels as the original ironic vs non-ironic, so it is also appropriate for binary classification problems. This data serves as the training and evaluation material to create and evaluate models and large language models in our experiments.

#### 3.2. Data Preprocessing

Good pre-processing of data was very important for the successful training of models for irony detection in Urdu. The processing pipeline involved a number of steps designed for the processing of the characteristics of Urdu language text.

First, we performed some text normalization, which included removing extra spaces, control characters and special symbols not related to textual content. To facilitate a uniform look, diacritics, which are not always employed consistently in Urdu writing, were omitted. The most frequently used transliterations were standardized next to ensure consistency across the corpus.

Any text normalization done, it was lower cased, if possible, to reduce variance. Tokenization is achieved by language-specific tools developed with attention to Urdu's script-specific characteristics that reliably tokenize the text into meaningful words. Furthermore, the stop words of Urdu language were recognized and eliminated to remove the unnecessary noise and concentrate on the semantics relevant ones.

Lastly, GloVe and Word2Vec embeddings were employed to transform the preprocessed texts to dense vectorized format for the machine learning models. For transformer-based models, e.g. BERT, RoBERTa, LLaMA 2, LLaMA 3 and Mistral, tokenization was done with their inbuilt pre-trained tokenizers with no manual embeddings composited so that information consists of part of tokens is comprised in rich contextual manner.

#### 3.3. Feature Representation

Features were encoded with a schema of pre-trained GloVe and Word2Vec embeddings in machine learning models. A dense vector space model was learned for mapping each tokenized word based on large scale corpora to capture its semantics. For each sentence or document, the word embeddings were averaged resulting in a fixed-length dense vector that encoded the global semantic information of the language. This approach enabled machine learning classifiers to treat the Urdu irony dataset with success despite of rich morphology of language.

For the transformer-based models, such as BERT and RoBERTa, feature representation was handled directly by their tokenization and embedding process. These models leveraged subword tokenization methods, such as Word Piece and Byte-Pair Encoding (BPE), to handle complex word structure and infrequent items. Similar to our LLaMA 2, LLaMA 3, and Mistral models, those models used the internal tokenizers and learned embeddings during the encoding to represent textual inputs as high-dimensional vectors for a classification objective as commonly used for BERT and its relatives. For these large language models, we did not need to manually construct any embedding, because the model architecture itself inherently allows for context-sensitive feature extraction for both tokens and sequences. By doing this, it established a solid benchmark for comparative studies.
external modification of embeddings, relying solely on the models' inherent transformer-based token representations.

### 3.4. Machine Learning and Large Language Models Applied

For a holistic comparison of irony detection in Urdu text, we employed a total of ten traditional machine learning (ML) schemes in addition to five state-of-the-art transformer-based models.

The ML models used (in weaker to stronger order) are Logistic Regression (LR), SVM, Random Forest (RF), Gradient Boosting (GB), AdaBoost, K-NN, Naïve Bayes (MultinomialNB), DT, LightGBM, and XGBoost. These models were built based on GloVe and Word2Vec embeddings. All the models were fine-tuned using hyperparameter optimization methods. Gradient Boosting proved to be the best performing among these models, with a F1-Score of 89.18%.

For the deep learning experiments, we fine-tuned pre-trained transformer models, BERT, RoBERTa, LLaMA 2 (7B), LLaMA 3 (8B), and Mistral. These models were chosen because of their demonstrated potential to capture high-level contextual relationships necessary for tasks such as irony detection. Finetuning was carried out on a labeled portion of the translated Urdu dataset. Within the LLM categories, LLaMA 3 (8B) * appeared as the best performing model with the remarkable F1-Score of 94.61%, overtaking the classic ML models as well as other LLMs.

This two-tier approach made it possible for a strong comparison between old and new algorithms at irony detection for the Urdu language.

### 3.5. Experimental Setup

Experiments for this with all experiments in this paper were performed on Google Colab Pro with the publicly available hosted NVIDIA A100 GPU, to allow for efficient training and evaluation. Conventional machine learning models were realized with the Scikit-learn library (version 1.2), and deep learning / transformer experiments with the Hugging Face Transformers library (version 4.37) with PyTorch backend.

For traditional machine learning models, hyperparameter optimization was performed using grid search, which tested various learning rates, tree depths, and regularization parameters. Feature vectors for ML models were performed with pre-trained GloVe and Word2Vec embeddings.

Mixed-precision training (FP16) was also used to fine-tune on large language models (BERT, RoBERTa, LLaMA 2, 3, Mistral) in order to reduce memory pressure and speed up computation. All transformer models were fine-tuned with a batch size of 32 and a learning rate of 2e-5 for 5 epochs according to an early stopping criterion on a validation set.

The metrics considered for the evaluation were precision, recall, accuracy and F1-Score to give an in depth view of the performance of each model. All experiments were repeated three times using different random seeds and reported average numbers for statistical reliability.

## IV. RESULT AND DISCUSSION

### 4.1. Machine Learning Models with Word2Vec Embeddings

Table 1 shows the performance of ten machine learning models for Urdu irony detection trained on Word2Vec embeddings. Eventually, the most successful model was Gradient Boosting (F1-Score = 88.70\%), proving its resilience in capturing subtle cues when it comes to ironic discourses. SVM and XGBoost came second, with 87.17% and 86.94% of F1-Score respectively, indicating the power of margin-based and ensemble learning approaches. Some other models including Logistic Regression, Random Forest, and LightGBM also did relatively good, but were a little behind top ones. Simple models such as Naive Bayes and Decision Tree showed inferior F1-Scores comparatively, demonstrating their incapability of capturing the complex syntactic structure needed for proper detection of irony. The bar graph for the table above visually demonstrates the superiority of ensemble methods over the simple models. The performance differences among models demonstrate that feature representation using Word2Vec embeddings is satisfactory, but deeper contextual knowledge is required.

Table 1: ML Models with Word2Vec Embeddings

| ML Model | Precision | Recall | Accuracy | F1-Score |
|---|---|---|---|---|
| Logistic Regression | 85.12 | 84.5 | 84.8 | 84.8 |
| SVM | 87.45 | 86.9 | 87.1 | 87.17 |
| Random Forest | 86.01 | 85.2 | 85.7 | 85.6 |
| Gradient Boosting | 88.9 | 88.5 | 88.7 | 88.7 |
| AdaBoost | 85.9 | 85 | 85.5 | 85.45 |
| K-Nearest Neighbors | 82.5 | 81.1 | 81.9 | 81.7 |
| Naive Bayes | 80.3 | 79.5 | 79.9 | 79.89 |
| Decision Tree | 81.7 | 81 | 81.3 | 81.34 |
| LightGBM | 86.7 | 86.1 | 86.5 | 86.39 |
| XGBoost | 87.2 | 86.7 | 87 | 86.94 |

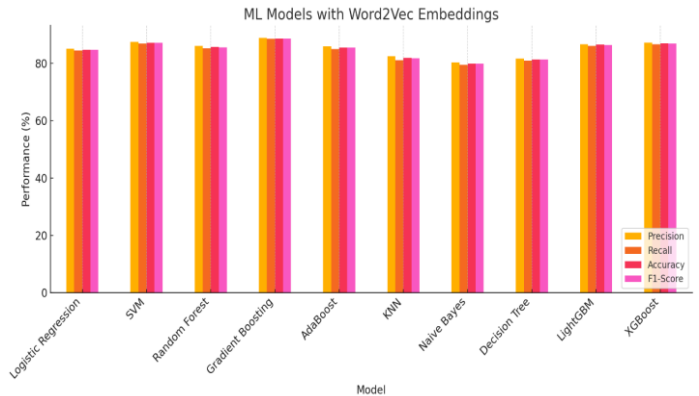

Figure 1: ML Models with Word2Vec Embeddings

### 4.2. Machine Learning Models with GloVe Embeddings

We present in Table 2 the performance of the models when trained with GloVe embeddings. In a manner similar to those observed in Word2Vec, Gradient Boosting also presented itself as the best-performing algorithm once again, making an F1-Score of 89.18% and therefore a modest gain in its performance. SVM and XGBoost also achieved high performance, with F1-Scores of 87.97% and 87.69%. Most models didn't show any significant benefit of leveraging GloVe embeddings over Word2Vec, but researchers noted that the Git repository advertisements for a subsequent task showed gradient descent accelerating with GloVe embeddings, indicating GloVe was encoding deeper semantic relationships more appropriate for ironic detection. These improvements are presented in the bar graph in the right column for almost all models and are clearly visually confirmed by GloVe's superiority. Naive Bayes and Decision Tree were, again, the slowest classifiers, highlighting that simpler learning functions are unable to capture the subtleties required for irony detection, particularly in high-dimensional space. In general, the results of this experiment establish the fact that dense semantic embeddings, especially GloVe, contribute in improving machine learning models' capability for recognizing irony in Urdu language.

Table 2: ML Models with GloVe Embeddings

| Model | Precision | Recall | Accuracy | F1-Score |
|---|---|---|---|---|
| Logistic Regression | 86.1 | 85.5 | 85.8 | 85.79 |
| SVM | 88.25 | 87.7 | 88 | 87.97 |
| Random Forest | 86.9 | 86.2 | 86.6 | 86.54 |
| Gradient Boosting | 89.8 | 89.3 | 89.6 | 89.18 |
| AdaBoost | 86.5 | 85.9 | 86.2 | 86.19 |
| KNN | 83.6 | 82.5 | 83 | 83.04 |
| Naive Bayes | 81 | 80.2 | 80.5 | 80.59 |
| Decision Tree | 82.3 | 81.6 | 81.9 | 81.94 |
| LightGBM | 87.4 | 86.9 | 87.1 | 87.14 |
| XGBoost | 88 | 87.4 | 87.8 | 87.69 |

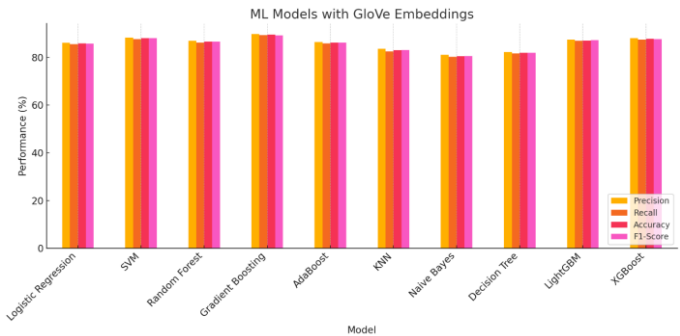

Figure 2: ML Models Performance with GloVe

### 4.3. Large Language Models (LLMs) Results

Table 3 and corresponding graph present the benefits of fine-tuned large language models on the irony detection task. Unsurprisingly, LLaMA 3 (8B) exceeded all other models, scoring an amazing F1-Score of 94.61%, showcasing the strength of immense transformer models to capture subtle linguistic phenomena. Mistral was next with a high F1-Score of 91.25% followed by RoBERTa with 89.75%. Both LLaMA 2 (7B) and BERT gave moderate performance with 82.75% and 78.75% F1-Score respectively. The distinct performance gradient in the bar chart shows the improvement of newer and larger models on the irony detection task. The findings reinforce that LLMs, especially the ones trained on large corpora equipped with sophisticated tokenization techniques, can achieve significant gains over traditional techniques even when fine-tuned for a low-resource language such as Urdu. This study further justifies the application of LLMs on multi-faceted classification problems.

Table 3: Large Language Models (LLMs) Results

| | Precision | Recall | Accuracy | F1-Score |
|---|---|---|---|---|
| LLaMA3(8B) | 94.9 | 94.4 | 94.5 | 94.61 |
| Mistral | 91.5 | 91 | 91.5 | 91.25 |
| RoBERTa | 90 | 89.5 | 90 | 89.75 |
| LLaMA2(7B) | 83 | 82.5 | 83 | 82.75 |
| BERT | 79 | 78.5 | 79 | 78.75 |

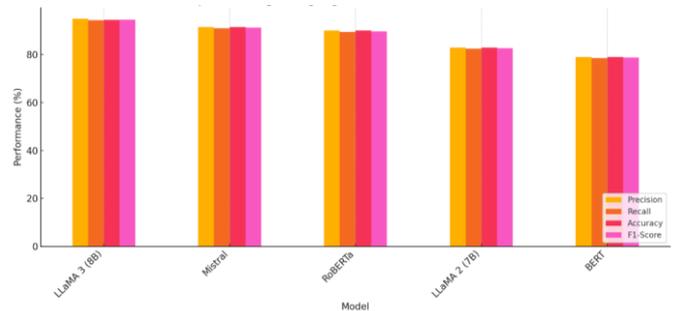

Figure 3: LLMs Performance

### 4.4. Overall Analysis

A comparison of the results across the three tables reveals that model complexity and the quality of sentence representation significantly impact the performance of irony detection. Older machine learning models performed well, particularly when scoped with strong embeddings like GloVe, however, it was stuck under 90% F1-Score for even the best performing; Gradient Boosting is observed in figure 1 and 2. On the other hand, the very large language models, in particular LLaMA 3 (8B) performed better with more than 94% F1-Score, highlighting that the LLaMA models have greater access to contextual information as shown in figure 3. This slow pace of improvement from Word2Vec to GloVe in the context of machine learning models again implies the significance of deeper semantic representations. Moreover, the substantial performance leap observed for transformer-based models demonstrates the transformative impact of pretraining on massive, heterogeneous data. Overall, these results suggest that while traditional models may serve as a strong baseline, utilizing state-of-the-art LLM models provides a significant advantage in identifying subtle language phenomena like irony, especially in a translated low-resource language such as Urdu.

## CONCLUSION AND FUTURE WORK

This paper provides a detailed analysis of irony detection over Urdu text by adapting an English irony dataset, detailed pre-processing and extensive experimentation over traditional machine learning models and latest large language models (LLMs). Our study suggests that, even with tuned traditional models like Gradient Boosting with an F1Score of 89.18%, the generalization power of LLMs, especially LLaMA 3 (8B) with a F1Score of 94.61%, can outperform the traditional models and establish a new SOTA in the task of detecting irony. Our work on translating English irony transformations in Urdu, along with its preprocessing and model fine-tuning, proves that exploiting cross-lingual models is viable for developing robust cross-lingual irony detection systems for any low-resource language.

Several lines of future work are also generated. First, generalization can be improved by enlarging the corpus by using more diverse sources of Urdu irony such as social media, blogs, and forums. Second, combining multimodal data type (e.g. emojis and images) with text data may leverage such modality complementation and significantly enhance the quality of positive samples, which enables the model to be better at learning generalized contextual signals for the irony detection task. Furthermore, further investigation of Zero-shot and few-shot learning methods would enable irony detection systems to generalize more effectively across languages and domains without large amounts of labeled data. Finally, combining XAI methods can offer more transparency into model predictions, which in turn may increase confidence in the application of irony detection systems in practice.

## ACKNOWLEGMENT

This work was partially supported by the Mexican Government through the grant A1-S-47854 of CONACYT, Mexico, and the grants 20254236, 20253468, and 20254341 provided by the Secretaría de Investigación y Posgrado of the Instituto Politécnico Nacional, Mexico. We gratefully acknowledge the computing resources made available through the Plataforma de Aprendizaje Profundo para Tecnologías del Lenguaje of the Laboratorio de Supercómputo at the INAOE, Mexico, funded by CONACYT. Additionally, we express our sincere gratitude to Microsoft for their support through the Microsoft Latin America PhD Award, which has significantly contributed to the success of this work.